\documentclass[10pt,twocolumn,letterpaper]{article}

\usepackage{cvpr}              

\definecolor{cvprblue}{rgb}{0.21,0.49,0.74}
\usepackage[pagebackref,breaklinks,colorlinks,allcolors=cvprblue]{hyperref}

\usepackage{multirow}

\def\paperID{*****} 
\def\confName{CVPR}
\def\confYear{2025}

\title{MedMoE: Modality-Specialized Mixture of Experts for Medical Vision-Language Understanding}

\author{Shivang Chopra\textsuperscript{1}, Gabriela Sanchez-Rodriguez\textsuperscript{1,2}, Lingchao Mao\textsuperscript{1}, Andrew J Feola\textsuperscript{1,2,3},  \\
Jing Li\textsuperscript{1}, Zsolt Kira\textsuperscript{1}\\
\textsuperscript{1}Georgia Institute of Technology \quad \textsuperscript{2}Emory University \quad \textsuperscript{3}Joseph M Cleland Atlanta VAMC\\
{\tt\small \{schopra47, lmao31, grodrguez3, jli3175, afeola3, zkira\}@gatech.edu}
}

\begin{document}

 \setlength{\abovedisplayskip}{5pt}
 \setlength{\belowdisplayskip}{5pt}

\maketitle
\begin{abstract}

Different medical imaging modalities capture diagnostic information at varying spatial resolutions, from coarse global patterns to fine-grained localized structures. However, most existing vision-language frameworks in the medical domain apply a uniform strategy for local feature extraction, overlooking the modality-specific demands. In this work, we present MedMoE, a modular and extensible vision-language processing framework that dynamically adapts visual representation based on the diagnostic context. MedMoE incorporates a Mixture-of-Experts (MoE) module conditioned on the report type, which routes multi-scale image features through specialized expert branches trained to capture modality-specific visual semantics. These experts operate over feature pyramids derived from a Swin Transformer backbone, enabling spatially adaptive attention to clinically relevant regions. This framework produces localized visual representations aligned with textual descriptions, without requiring modality-specific supervision at inference. Empirical results on diverse medical benchmarks demonstrate that MedMoE improves alignment and retrieval performance across imaging modalities, underscoring the value of modality-specialized visual representations in clinical vision-language systems. 
\end{abstract}    
\vspace{-10pt}
\section{Introduction}
\label{sec:intro}

Recent advances in vision language models (VLM) \cite{liu2023llava} have demonstrated great performance across a wide range of tasks such as zero-shot classification, segmentation, and visual question-answering. However, these models are less effective in medical domains, where data significantly differs from web content \cite{singhal2023large}. To bridge this gap, domain-specific medical VLMs have been developed by pretraining on radiology images paired with diagnostic reports or captions \cite{radford2021clip}.  
A key insight driving recent progress is the incorporation of local contrastive learning \cite{huang2021gloria} and multi-scale visual semantics \cite{zhou2023mrm} , which can significantly improve the fine-grained semantic alignment between regions in an image and words in a report \cite{huang2021gloria}.

\begin{figure}[t]
  \centering
  \includegraphics[width=0.9\linewidth]{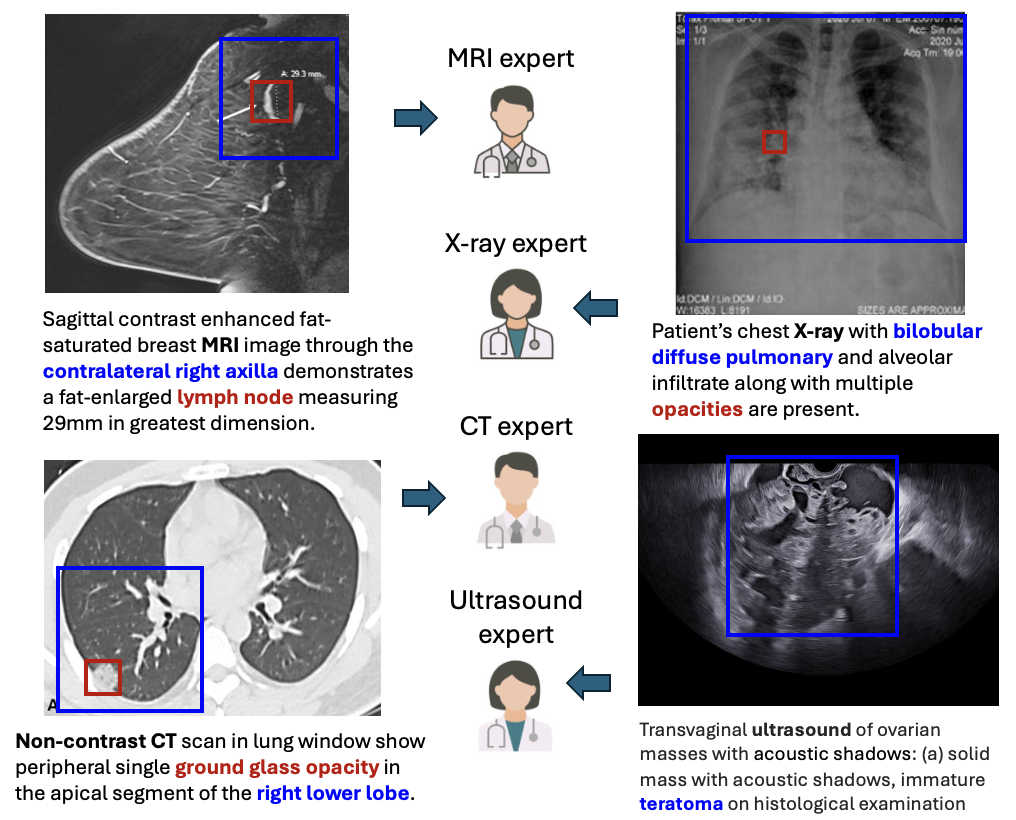}
  \caption{
  \textbf{Motivation: diagnostic context varies across modalities.}
  Each image-report pair exemplifies how clinical observations may focus on localized lesions (red) or broader regions (blue), depending on the imaging modality and diagnostic task. \vspace{-15pt}
  }
  \label{fig:motivation}
\end{figure}

However, current approaches typically employ a \emph{static} strategy for feature extraction, applying a uniform processing pipeline regardless of the imaging modality. This design overlooks the variations across imaging modalities, driven by inherent differences in image resolution and representation. As a result, the nature of word-region alignment can differ significantly. For example, MRI reports often emphasize local abnormalities, while interpretation of X-rays often focus on global patterns. Figure~\ref{fig:motivation} illustrates this variation, where clinical focus shifts between localized lesions and broader regions based on modality. Existing models trained predominantly on datasets like chest X-rays struggle to generalize to more diverse settings \cite{zhao2025clipsurvey}, and even models trained on broader datasets such as UniMed \cite{khattak2024unimed} employ a single shared encoder for all modalities, potentially limiting their adaptability to heterogeneous visual cues.

To overcome these limitations, we propose \textbf{MedMoE}, a modular vision-language framework that introduces Mixture-of-Experts (MoE) into the local visual processing pipeline to enable conditional specialization based on the diagnostic context. MoE-based architectures have gained prominence in large-scale language models \cite{dai2024deepseekmoeultimateexpertspecialization} and multimodal systems \cite{lin2024moe, wu2024deepseekvl2mixtureofexpertsvisionlanguagemodels}, where they improve efficiency and generalization by dynamically routing inputs through specialized expert modules. MedMoE routes the local visual features extracted from the image through a set of modality-aware expert branches. Each expert is trained to specialize in a distinct diagnostic context, allowing the model to selectively emphasize spatial features aligned with the semantic granularity of the associated report. This dynamic routing mechanism enables MedMoE to adapt visual grounding to modality-specific needs, outperforming existing multi-scale contrastive learning methods across a range of medical benchmarks. These findings highlight the promise of adaptive expert-driven architectures in building robust and scalable medical VLMs.

\section{Related Work}
\label{sec:relwork}

\paragraph{Global and Local Representation Learning.}
Early medical VLMs such as MedCLIP~\cite{wang-etal-2022-medclip} focus primarily on global image-text alignment using contrastive learning. These models treat the entire report as a single semantic unit and align it with a global image embedding, limiting their capacity to capture fine-grained regional semantics critical in diagnostic tasks. GLoRIA~\cite{huang2021gloria} introduced a global-local contrastive objective, allowing individual words in the report to attend to semantically relevant subregions of the image. Subsequent works such as LoVT~\cite{mueller2022lovt} and PRIOR~\cite{cheng2023prior} further enhanced this alignment using sentence-level supervision and attention weighting mechanisms. However, these methods operate with a static, modality-agnostic encoder, which restricts their ability to adapt local grounding strategies to the diagnostic granularity required across diverse imaging modalities. In contrast, \textbf{MedMoE} enables context-aware specialization by routing multi-scale visual features through expert branches conditioned on the report content, allowing for adaptive alignment tailored to both the report semantics and imaging type.

\paragraph{Multi-Scale Medical Vision-Language Pretraining.}
Multi-scale representation learning has emerged as a key challenge for adapting CLIP-style models to the medical domain~\cite{zhao2025clipsurvey}, where clinically salient features can range from global abnormalities (e.g., X-rays) to focal lesions (e.g., CT, MRI). Recent approaches like MRM~\cite{zhou2023mrm} and fVLM~\cite{shui2025fv} address this by incorporating architectural changes such as hierarchical decoders, reconstruction losses, or anatomy-guided alignment. While these methods improve scale-aware learning, they apply a fixed processing pipeline and shared encoder across all modalities, thus failing to account for the heterogeneity in spatial detail and diagnostic focus. \textbf{MedMoE} addresses this limitation by enabling dynamic feature specialization through expert pathways, where routing is informed by modality-specific report context.

\paragraph{Mixture-of-Experts and Context-Aware Specialization.}
MoE architectures have proven effective for scaling language models using sparse conditional computation~\cite{shazeer2017outrageously,lepikhin2020gshard}, and recent work has extended MoE to multimodal VLMs for efficient and specialized fusion~\cite{du2022glam,zhou2022mixture}. However, their application to medical vision-language learning remains limited. Existing VLMs typically apply MoE at the fusion or output stage, with limited impact on early visual representation learning. We propose \textbf{MedMoE}, the first medical VLM to integrate report-conditioned MoE routing within the visual feature extractor itself. This design enables modality- and task-specific expert selection for adaptive local grounding, addressing the bottleneck of static, one-size-fits-all processing in existing pipelines.

\section{Method}
\label{sec:method}

\section{Methodology}

\begin{figure*}
    \centering
    \includegraphics[width=0.85\linewidth]{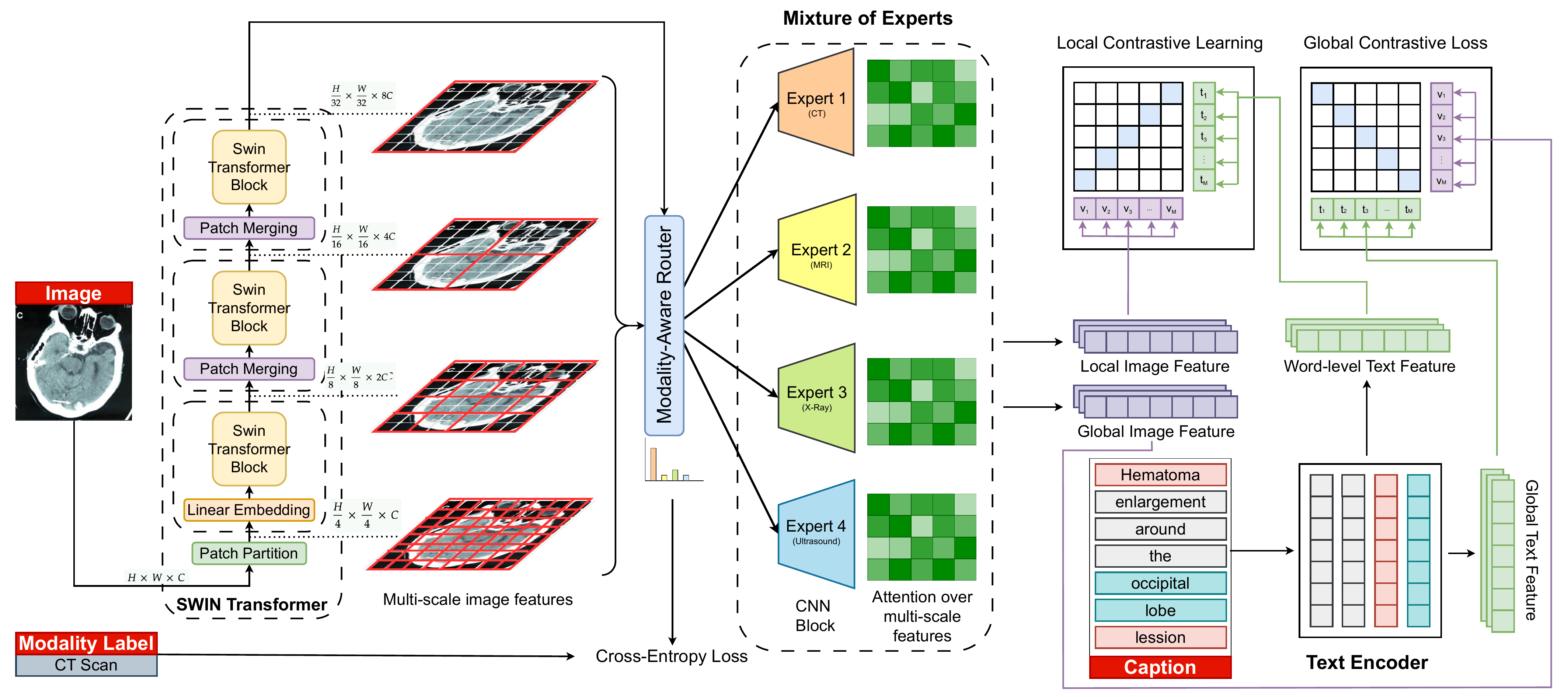} 
    \vspace{-10pt}
    \caption{\textbf{Overall architecture of MedMoE.} A Swin Transformer extracts multi-scale features, and a modality-aware router selects a specialized expert based on the global image embedding. The expert outputs local embeddings aligned with word-level text via a local contrastive loss. Global contrastive loss and auxiliary loss enforce image-text alignment and supervise expert selection respectively.\vspace{-10pt}}
    \label{fig:pipeline}
\end{figure*}

We propose \textbf{MedMoE}, a diagnostic-aware Mixture-of-Experts (MoE) framework for context-sensitive local feature extraction in medical vision-language modeling. MedMoE extends the global-local contrastive paradigm established by GLoRIA~\cite{huang2021gloria}, while introducing a dynamic, report-conditioned mechanism for multi-scale local visual representation. Specifically, MedMoE replaces static feature extractors with an MoE module that routes multi-scale visual features through modality-specialized expert branches. The routing is conditioned on the diagnostic report, enabling adaptive specialization. The overall schematics of MedMoE are illustrated in Figure \ref{fig:pipeline}.

\subsection{Problem Setup}

Let $(\mathbf{x}_v, \mathbf{x}_t)$ denote a paired medical image and its diagnostic report.  
From each pair, we extract a global image embedding $\mathbf{v}_g \in \mathbb{R}^D$, a local image‑embedding grid $\mathbf{V} \in \mathbb{R}^{M \times D}$, a global text embedding $\mathbf{t}_g \in \mathbb{R}^D$, and token‑level text embeddings $\mathbf{T}=[\mathbf{t}_1,\dots,\mathbf{t}_N] \in \mathbb{R}^{N \times D}$.  
Our objective is to learn semantically aligned representations that maximize image-text alignment at both global and token-region levels.

\subsection{Multi-Scale Visual Feature Extraction}

We adopt the Swin Transformer~\cite{liu2021swin} as our image encoder backbone due to its hierarchical design and strong spatial sensitivity. The encoder outputs a set of multi-scale features:
\[
\mathcal{F} = \{ \mathbf{F}^{(1)}, \dots, \mathbf{F}^{(L)} \},
\]
where each $\mathbf{F}^{(l)}$ corresponds to feature maps of spatial resolution $\{\frac{H}{4}{\times}\frac{W}{4}, \frac{H}{8}{\times}\frac{W}{8}, \frac{H}{16}{\times}\frac{W}{16}, \frac{H}{32}{\times}\frac{W}{32}\}$, respectively. These feature maps form the input to our expert-specific local feature heads.

\subsection{Report-Conditioned Mixture-of-Experts}

To support diagnostic-aware specialization, we introduce a Mixture-of-Experts (MoE) module composed of $K$ convolutional experts $\{E_k\}_{k=1}^K$, each trained to focus on distinct diagnostic modalities and learn modality-specific spatial reasoning patterns.

Each expert processes a set of hierarchical multi-scale features $\mathcal{F} = \{ \mathbf{F}^{(1)}, \dots, \mathbf{F}^{(L)} \}$ extracted from different stages of the Swin Transformer. To produce a unified local representation, each expert first projects features at each scale to a shared embedding space and aligns their spatial resolutions via interpolation. Rather than uniformly aggregating across scales, we introduce an \textit{cross-scale attention mechanism} that dynamically weights the contribution of each scale. For a given expert $E_k$, the fused local representation is computed as:
\[
\mathbf{V}_k = \sum_{\ell=1}^{L} \beta^{(\ell)}_{k} \cdot \mathbf{F}^{(\ell)}_k,
\]
where $\beta^{(\ell)}_{k}$ are soft attention weights predicted at each spatial location by the expert's scale-attention module. This allows each expert to adaptively combine fine and coarse features depending on the image content and diagnostic specialization.

To ensure computational efficiency and promote specialization, we use a \textit{hard routing} strategy to select a single expert per input. The routing module computes scores $\boldsymbol{\alpha} = [\alpha_1, \dots, \alpha_K]$ via a lightweight MLP conditioned on the global report embedding $\mathbf{t}_g$:

\[
\boldsymbol{\alpha} = \text{softmax}\left( W_2 \cdot \text{ReLU}(W_1 \cdot \mathbf{t}_g) \right),
\]
with $W_1$ and $W_2$ as learned parameters. The index of the selected expert is then given by $k^* = \arg\max_k \alpha_k$. The final local visual representation is defined as:
\[
\mathbf{V}_{\text{local}} = \mathbf{V}_k*
\]
This hard selection mechanism avoids computing all expert branches during inference, thereby reducing overhead while still leveraging diagnostic-aware specialization. This two-level specialization, across both scales and experts, enables the model to flexibly adapt to diverse diagnostic imaging contexts.

\subsection{Contrastive Learning Objectives}
Following GLoRIA~\cite{huang2021gloria}, we apply both global and local contrastive learning for image-text alignment: \\
\textbf{Global Contrastive Loss.} The global loss aligns the image-level and report-level representations:
\[
\mathcal{L}_{\text{global}} = -\log \frac{\exp(\text{sim}(\mathbf{v}_g, \mathbf{t}_g)/\tau)}{\sum_{j=1}^{B} \exp(\text{sim}(\mathbf{v}_g, \mathbf{t}_j)/\tau)},
\]
where $\tau$ is the temperature hyperparameter, $B$ is the batch size, and $\text{sim}(\cdot, \cdot)$ denotes cosine similarity. \\
\textbf{Local Contrastive Loss.} For each token $\mathbf{t}_i$, we compute a visual context vector $\mathbf{c}_i$ by attending to local features $\mathbf{v}_j$:
\[
a_{ij} = \frac{\exp(\mathbf{t}_i^\top \mathbf{v}_j / \tau)}{\sum_j^M \exp(\mathbf{t}_i^\top \mathbf{v}_j / \tau)}, \quad
\mathbf{c}_i = \sum_j^M a_{ij} \cdot \mathbf{v}_j,
\]
\[
\mathcal{L}_{\text{local}} = \sum_i^N -\log \frac{\exp(\text{sim}(\mathbf{c}_i, \mathbf{t}_i)/\tau)}{\sum_{j}^N \exp(\text{sim}(\mathbf{c}_i, \mathbf{t}_j)/\tau)}.
\]

\begin{table*}
    \centering
    \resizebox{\textwidth}{!}{ 
    \begin{tabular}{lcccccccccccccc}
        \toprule
        \multirow{2}{*}{\textbf{Approach}} & \multirow{2}{*}{\textbf{Dataset}} & \multirow{2}{*}{\textbf{Modality}} & \multirow{2}{*}{\textbf{Image Encoder}} & \multicolumn{2}{c}{\textbf{X-ray}} & \multicolumn{2}{c}{\textbf{Ultrasound}} & \multicolumn{2}{c}{\textbf{MRI}} & \multicolumn{3}{c}{\textbf{CT}} & \multirow{2}{*}{\textbf{Avg.}} \\
        \cmidrule(lr){5-6} \cmidrule(lr){7-8} \cmidrule(lr){9-10} \cmidrule(lr){11-13} 
        & & & & \textbf{CheXpert(5x200)} & \textbf{RSNA} & \textbf{Thyroid} & \textbf{Breast} & \textbf{ACL} & \textbf{Meniscus} & \textbf{Axial} & \textbf{Coronal} & \textbf{Sagittal} & \\
        \midrule
         \multicolumn{13}{c}{\textbf{Specialist Models}} \\
         \midrule
        QuiltNet \cite{park2022quiltnet} & Quilt (1M) \cite{ikezogwo2023quilt} & Histopathology & ViT-B/16 & 18.20 & 45.45 & 60.12 & 42.27 & 34.49 & 55.57 & 10.03 & 5.83 & 9.38 & 31.26 \\
        KeepFIT \cite{wu2024mm} & MM-Retinal \cite{wu2024mm} & Fundus & ResNet50 & 23.30 & 50.00 & 40.12 & 59.72 & 66.32 & 62.26 & 8.74 & 12.46 & 9.06 & 36.88 \\
        GLoRIA \cite{huang2021gloria}  & CheXpert (200k) \cite{irvin2019chexpert} & X-Ray  & ResNet50 & 61.00 & 50.00 & 79.16 & 72.43 & 43.75 & \underline{76.44} & 3.39 & 3.39 & 3.39 & 43.66 \\
        MedCLIP \cite{wang-etal-2022-medclip} & CheXpert (200k) \cite{irvin2019chexpert} & X-Ray  & SWIN & 59.42 & 73.63 & 44.68 & 53.98 & 68.01 & 45.27 & 9.87 & 13.92 & 10.36 & 42.12 \\
        \midrule
        \multicolumn{13}{c}{\textbf{Generalist Models}} \\
        \midrule
        PMC-CLIP \cite{lin2023pmc} & PMC-OA (1.5M) \cite{lin2023pmc} & All & ResNet50 & 30.30 & \underline{74.99} & 58.60 & \underline{68.76} & 67.09 & 65.03 & 33.98 & 20.23 & 23.30 & 49.14 \\
        BiomedCLIP \cite{zhang2023biomedclip} & PMC (15M) \cite{zhang2023biomedclip} & All & ViT-B/16 & 38.30 & 72.56 & 61.05 & 68.12 & 47.89 & 40.09 & 29.13 & 19.09 & 20.39 & 44.06\\
        UniMed-CLIP \cite{khattak2024unimed} & UniMed (5.3M) \cite{khattak2024unimed} & All & ViT-B/16 & \underline{65.90} & 71.65 & \textbf{71.35} & 66.31 & \textbf{85.82} & \textbf{85.27} & \underline{37.54} & \underline{24.43} & \underline{31.72} & \textbf{59.98} \\
        \midrule
      \rowcolor{gray!20} Ours & UniMed (5.3M) \cite{khattak2024unimed} & All & SWIN + MoE & \textbf{66.03} & \textbf{78.32} & \underline{62.74} & \textbf{69.32} & \underline{77.92} & 74.87 & \textbf{40.00} & \textbf{28.32} & \textbf{26.83} & \underline{58.26} \\
        \bottomrule
    \end{tabular}
    }
    \vspace{-10pt}
    \caption{\textbf{Zero-shot classification accuracy on various radiology datasets.} Best results are highlighted in bold and second-best results are underlined. \vspace{-15pt}}
    \label{tab:radiology-results} 
\end{table*}

\subsection{Auxiliary Report-Type Supervision}

To reinforce the specialization of experts, we introduce an auxiliary classification head over the report embedding:
\[
\mathcal{L}_{\text{aux}} = \text{CrossEntropy}(W_c \cdot \mathbf{t}_g, y),
\]
where $W_c$ is a classification head and $y$ is the ground-truth report type (e.g., CT, X-ray, MRI). This encourages the router to learn modality-sensitive routing that aligns with variations in real diagnostic contexts.

\subsection{Overall Training Objective}

The final training loss is a weighted combination of the above objectives:
\[
\mathcal{L}_{\text{total}} = \mathcal{L}_{\text{global}} + \mathcal{L}_{\text{local}} + \lambda \cdot \mathcal{L}_{\text{aux}},
\]
where $\lambda$ is a tunable hyperparameter to balance the auxiliary supervision.

\begin{table}[h]
\centering
\resizebox{\linewidth}{!}{
\begin{tabular}{c|c|c c c}
\hline
\textbf{Dataset} & \multirow{2}{*}{\textbf{Model}} & \multicolumn{3}{c}{\textbf{Performance \%}} \\
\cline{3-5} 
(Modality) &  & \textbf{1\%} & \textbf{10\%} & \textbf{100\%} \\ \hline \hline
\multirow{5}{*}{RSNA \cite{colak2021rsna}} & CLIP \cite{radford2021clip} & 71.67 & 73.89 & 81.09 \\ 
                      & PMC-CLIP \cite{lin2023pmc} & 65.25 & 64.15 & 79.47 \\ 
                      & BiomedCLIP \cite{zhang2023biomedclip} & \underline{80.04} & 78.32 & 83.84 \\ 
                    (X-ray)  & UniMed-CLIP \cite{khattak2024unimed} &  79.52 &  \underline{79.19} &  \textbf{88.51} \\
     & \cellcolor{gray!20} Ours & \cellcolor{gray!15} \textbf{81.12} & \cellcolor{gray!15} \textbf{83.84} & \cellcolor{gray!15} \underline{87.83} \\ \hline
\multirow{5}{*}{Thyroid \cite{pedraza2015open} } & CLIP \cite{radford2021clip} & 70.99 & 72.51 & 75.99 \\ 
                          & PMC-CLIP \cite{lin2023pmc} & 46.78 & 56.61 & 61.40 \\ 
                          & BiomedCLIP \cite{zhang2023biomedclip} & \textbf{76.96} & \textbf{80.47} & \textbf{81.87} \\ 
                        (Ultrasound)  &  UniMed-CLIP \cite{khattak2024unimed}  &  55.67 &  64.44 & 76.84 \\
     & \cellcolor{gray!20} Ours & \cellcolor{gray!20} \underline{73.99} & \cellcolor{gray!20} \underline{76.96} & \cellcolor{gray!20} \underline{79.83} \\ \hline
\multirow{5}{*}{ACL } & CLIP \cite{radford2021clip} & 56.99 & 85.89 & 91.73 \\ 
                      & PMC-CLIP \cite{lin2023pmc} & 57.00 & 63.47 & 77.92 \\ 
                      & BiomedCLIP \cite{zhang2023biomedclip} & 44.13 & 65.46 & 90.12 \\ 
                    (MRI)   & UniMed-CLIP \cite{khattak2024unimed} & \textbf{89.61} &  \textbf{95.28} & \textbf{97.28} \\
     & \cellcolor{gray!20} Ours & \cellcolor{gray!20} \underline{85.89} & \cellcolor{gray!20} \underline{89.77} & \cellcolor{gray!20} \underline{92.84} \\ \hline
\multirow{5}{*}{MediMeTA Axial} & CLIP \cite{radford2021clip} & 32.20 & 45.97 & 70.06 \\ 
                                & PMC-CLIP \cite{lin2023pmc} & 35.92 & 43.04 & 57.61 \\ 
                                & BiomedCLIP \cite{zhang2023biomedclip} & 29.77 & \underline{59.06} & \textbf{77.67} \\ 
                              (CT)  & UniMed-CLIP \cite{khattak2024unimed} & \underline{39.97} &  57.28 & 76.38 \\
     & \cellcolor{gray!20} Ours & \cellcolor{gray!20} \textbf{42.68} & \cellcolor{gray!20} \textbf{62.13} & \cellcolor{gray!20} \underline{76.38} \\ \hline
\end{tabular}
}
\vspace{-10pt}
\caption{\textbf{Linear Probing Results.} Performance (Accuracy) comparison across different generalist medical foundation models with varying training data percentages. Best results are highlighted in bold and second-best results are underlined. \vspace{-10pt}}
\label{tab:linear_probing_table}
\end{table}

\section{Experiments and Results}
\label{sec:experiments}

We evaluate MedMoE in terms of zero-shot and linear probing classification performance across diverse medical imaging modalities and benchmarks, demonstrating its effectiveness in context-sensitive visual-textual alignment.

\subsection{Experimental Setup}
\textbf{Datasets:} We use the UniMed \cite{khattak2024unimed} dataset for pretraining and evaluate on diagnostic benchmarks used in previous works (e.g., RSNA \cite{colak2021rsna}, CheXpert \cite{irvin2019chexpert}, ACL, Meniscus, Thyroid \cite{pedraza2015open}, Breast) covering four modalities: X-ray, MRI, CT, and Ultrasound. For linear probing, we use 1\%, 10\%, and 100\% of the training set from each benchmark.

\noindent \textbf{Baselines:} We compare MedMoE with generalist VLMs (CLIP \cite{radford2021clip}, PMC-CLIP \cite{lin2023pmc}, BiomedCLIP \cite{zhang2023biomedclip}, UniMed-CLIP \cite{khattak2024unimed}) and specialist models (MedCLIP \cite{wang-etal-2022-medclip}, QuiltNet \cite{park2022quiltnet}, MM-Retinal \cite{wu2024mm}). 

\noindent \textbf{Implementation Details:} We initialize the Swin Transformer \cite{liu2021swin} backbone from MedCLIP-pretrained weights \cite{deng2009imagenet}. The expert branches are 3-layer convolutional modules with BatchNorm and ReLU activations. Training is conducted on 8 NVIDIA A40 GPUs with a global batch size of 256, using gradient accumulation with 10 steps.

\subsection{Zero‑Shot Classification}
Table~\ref{tab:radiology-results} reports zero-shot accuracies on nine radiology benchmarks spanning four imaging modalities \textbf{MedMoE attains the best performance on 6/9 datasets}, including \textbf{78.32\%} on RSNA and \textbf{69.32\%} on Breast Ultrasound, outperforming the strongest generalist baseline (PMC‑CLIP) by 3.33\% and 0.56\%, respectively.  
On MRI datasets, MedMoE achieves \textbf{77.92\%} on ACL and \textbf{74.87\%} on Meniscus, an improvement of 30.0\% and 34.8\% over BioMedCLIP, while remaining competitive with UniMed‑CLIP, which is tuned specifically for MRI. Finally, MedMoE delivers state‑of‑the‑art results on all three CT views (40.00\%, 28.32\%, 26.83\%).

\begin{figure*}
    \centering
    \includegraphics[width=0.7\linewidth]{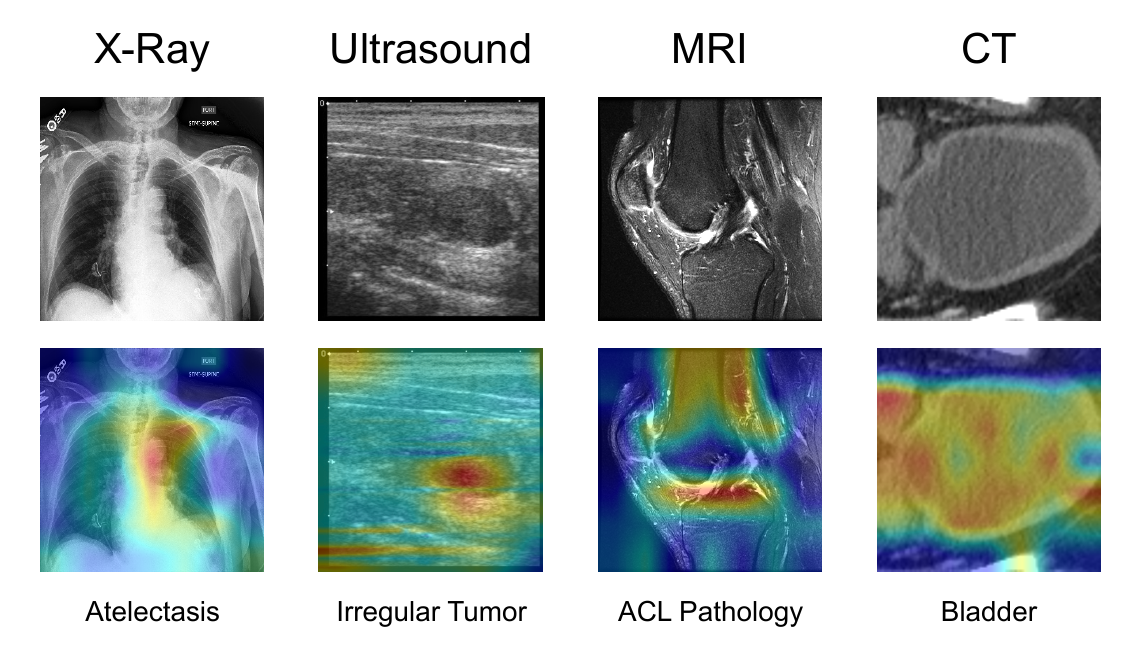}
    \vspace{-10pt}
    \caption{\textbf{Visualization of Attention Weights across Modalities.} Word-level attention maps are shown for different imaging modalities (X-ray, Ultrasound, MRI, and CT), highlighting regions relevant to the associated diagnostic labels. \vspace{-5pt}}
    \label{fig:attention_vis}
\end{figure*}

\subsection{Linear Probing Transfer}
We next freeze the visual encoder and fit a single linear layer on 1\%, 10\%, and 100\% of each training set (Table~\ref{tab:linear_probing_table}).  
In the extreme low‑data regime (1\%), MedMoE yields \textbf{81.12\%} on RSNA, surpassing the previous best BiomedCLIP (80.04\%). On Thyroid Ultrasound, MedMoE reaches \textbf{73.99\%}, a gain of 27.21\% over PMC‑CLIP, and only 2.97\% behind the modality‑specialised BiomedCLIP. For MRI (ACL) and CT (MediMeTA Axial), MedMoE achieves \textbf{85.89\%} and \textbf{42.68\%}, respectively, matching or exceeding all generalist baselines. 

\section{Discussion}
\label{sec:discussion}

\subsection{Visualization of Attention Weights}
Figure~\ref{fig:attention_vis} illustrates the attention maps produced by MedMoE’s expert branches across four imaging modalities. Each expert attends to diagnostically relevant regions specific to its modality: lung outlines in X-rays for detecting atelectasis, heterogeneous thyroid tissue in ultrasound, ligament structures in MRI, and abdominal organs in CT. These visualizations highlight MedMoE’s ability to perform modality-specific visual grounding. By routing features through specialized experts, the model learns to adapt its spatial focus to the granularity and diagnostic patterns unique to each modality.

\subsection{Computational Cost Comparison}

While MedMoE introduces multiple expert branches, it uses \textit{hard expert selection} at inference time, activating only a single expert per input. As a result, its computational cost remains comparable to baseline models built on the Swin-Tiny backbone. Table~\ref{tab:flops} reports the FLOPs and parameter counts of MedMoE alongside representative medical vision-language models. Although all experts are included during training, only one is active at inference, resulting in minimal overhead. This design enables MedMoE to achieve modality-aware specialization while maintaining efficiency and scalability.

\begin{table}[h]
\centering

\resizebox{\linewidth}{!}{
\begin{tabular}{lccc}
\toprule
\textbf{Model} & \textbf{Backbone} & \textbf{Params (M)} & \textbf{FLOPs (G)} \\
\midrule
CLIP (RN50)    & ResNet-50              & 38.3   & 6.12  \\
BiomedCLIP     & ViT-B/16               & 86.1   & 16.8 \\
UniMed-CLIP    & ViT-B/16               & 86.1   & 11.29 \\
MedCLIP        & Swin-Tiny              & 27   & 4.5  \\
MedMoE (Ours)  & Swin-Tiny + MoE  & 37\textsuperscript{*} & 7.8\textsuperscript{*} \\
\bottomrule
\end{tabular}
}
\vspace{0.5em}
\caption{\textbf{FLOPs and parameter comparison.} MedMoE uses Swin-T and activates only one expert at inference. *MedMoE parameter count includes all experts; FLOPs are measured with a single expert active.}
\label{tab:flops}
\end{table}
\section{Conclusion and Future Work}
\label{sec:conclusion}

We present \textbf{MedMoE}, a vision–language model that dynamically routes multi-scale visual features through a diagnosis-conditioned Mixture-of-Experts framework. By leveraging report-aware specialization, MedMoE achieves state-of-the-art performance on both zero-shot and linear probing benchmarks across diverse medical imaging modalities. In addition to strong quantitative gains, we provide qualitative analyses that illustrate improved modality-sensitive visual grounding, and computational comparisons that demonstrate MedMoE’s efficiency through hard expert selection. These results underscore the importance of modality-specific feature routing and context-aware alignment in medical vision-language learning. Future work will explore extending MedMoE to a wider range of unimodal and multimodal tasks, such as segmentation, report generation, and few-shot adaptation.
{
    \small
    \bibliographystyle{ieeenat_fullname}
    \bibliography{main}

\begin{thebibliography}{28}
\providecommand{\natexlab}[1]{#1}
\providecommand{\url}[1]{\texttt{#1}}
\expandafter\ifx\csname urlstyle\endcsname\relax
  \providecommand{\doi}[1]{doi: #1}\else
  \providecommand{\doi}{doi: \begingroup \urlstyle{rm}\Url}\fi

\bibitem[Cheng and et~al.(2023)]{cheng2023prior}
Bowen Cheng and et al.
\newblock Prior: Prototype-driven radiograph interpretation with optimized representations.
\newblock In \emph{NeurIPS}, 2023.

\bibitem[Colak et~al.(2021)Colak, Kitamura, Hobbs, Wu, Lungren, Prevedello, Kalpathy-Cramer, Ball, Shih, Stein, et~al.]{colak2021rsna}
Errol Colak, Felipe~C Kitamura, Stephen~B Hobbs, Carol~C Wu, Matthew~P Lungren, Luciano~M Prevedello, Jayashree Kalpathy-Cramer, Robyn~L Ball, George Shih, Anouk Stein, et~al.
\newblock The rsna pulmonary embolism ct dataset.
\newblock \emph{Radiology: Artificial Intelligence}, 3\penalty0 (2):\penalty0 e200254, 2021.

\bibitem[Dai et~al.(2024)Dai, Deng, Zhao, Xu, Gao, Chen, Li, Zeng, Yu, Wu, Xie, Li, Huang, Luo, Ruan, Sui, and Liang]{dai2024deepseekmoeultimateexpertspecialization}
Damai Dai, Chengqi Deng, Chenggang Zhao, R.~X. Xu, Huazuo Gao, Deli Chen, Jiashi Li, Wangding Zeng, Xingkai Yu, Y. Wu, Zhenda Xie, Y.~K. Li, Panpan Huang, Fuli Luo, Chong Ruan, Zhifang Sui, and Wenfeng Liang.
\newblock Deepseekmoe: Towards ultimate expert specialization in mixture-of-experts language models, 2024.

\bibitem[Deng et~al.(2009)Deng, Dong, Socher, Li, Li, and Fei-Fei]{deng2009imagenet}
Jia Deng, Wei Dong, Richard Socher, Li-Jia Li, Kai Li, and Li Fei-Fei.
\newblock Imagenet: A large-scale hierarchical image database.
\newblock In \emph{2009 IEEE Conference on Computer Vision and Pattern Recognition}, pages 248--255. IEEE, 2009.

\bibitem[Du(2022)]{du2022glam}
Yuxin et~al. Du.
\newblock Glam: Efficient scaling of language models with mixture-of-experts.
\newblock In \emph{ICML}, 2022.

\bibitem[Huang et~al.(2021)Huang, Shen, Lungren, and Yeung]{huang2021gloria}
Shih-Cheng Huang, Liyue Shen, Matthew~P. Lungren, and Serena Yeung.
\newblock Gloria: A multimodal global-local representation learning framework for label-efficient medical image recognition.
\newblock In \emph{Proceedings of the IEEE/CVF International Conference on Computer Vision (ICCV)}, 2021.

\bibitem[Ikezogwo et~al.(2023)Ikezogwo, Seyfioglu, Ghezloo, Geva, Sheikh~Mohammed, Anand, Krishna, and Shapiro]{ikezogwo2023quilt}
Wisdom Ikezogwo, Saygin Seyfioglu, Fatemeh Ghezloo, Dylan Geva, Fatwir Sheikh~Mohammed, Pavan~Kumar Anand, Ranjay Krishna, and Linda Shapiro.
\newblock Quilt-1m: One million image-text pairs for histopathology.
\newblock \emph{Advances in neural information processing systems}, 36:\penalty0 37995--38017, 2023.

\bibitem[Irvin et~al.(2019)Irvin, Rajpurkar, Ko, Yu, Ciurea-Ilcus, Chute, Marklund, Haghgoo, Ball, Shpanskaya, et~al.]{irvin2019chexpert}
Jeremy Irvin, Pranav Rajpurkar, Michael Ko, Yifan Yu, Silviana Ciurea-Ilcus, Chris Chute, Henrik Marklund, Behzad Haghgoo, Robyn Ball, Katie Shpanskaya, et~al.
\newblock Chexpert: A large chest radiograph dataset with uncertainty labels and expert comparison.
\newblock In \emph{Proceedings of the AAAI Conference on Artificial Intelligence}, pages 590--597, 2019.

\bibitem[Khattak et~al.(2024)Khattak, Kunhimon, Naseer, Khan, and Khan]{khattak2024unimed}
Muhammad~Uzair Khattak, Shahina Kunhimon, Muzammal Naseer, Salman Khan, and Fahad~Shahbaz Khan.
\newblock Unimed-clip: Towards a unified image-text pretraining paradigm for diverse medical imaging modalities.
\newblock \emph{arXiv preprint arXiv:2412.10372}, 2024.

\bibitem[Lepikhin(2020)]{lepikhin2020gshard}
Dmitry et~al. Lepikhin.
\newblock Gshard: Scaling giant models with conditional computation and automatic sharding.
\newblock In \emph{ICLR}, 2020.

\bibitem[Lin et~al.(2024)Lin, Tang, Ye, Cui, Zhu, Jin, Zhang, Ning, and Yuan]{lin2024moe}
Bin Lin, Zhenyu Tang, Yang Ye, Jiaxi Cui, Bin Zhu, Peng Jin, Junwu Zhang, Munan Ning, and Li Yuan.
\newblock Moe-llava: Mixture of experts for large vision-language models.
\newblock \emph{arXiv preprint arXiv:2401.15947}, 2024.

\bibitem[Lin et~al.(2023)Lin, Zhao, Zhang, Wu, Zhang, Wang, and Xie]{lin2023pmc}
Weixiong Lin, Ziheng Zhao, Xiaoman Zhang, Chaoyi Wu, Ya Zhang, Yanfeng Wang, and Weidi Xie.
\newblock Pmc-clip: Contrastive language-image pre-training using biomedical documents.
\newblock In \emph{International Conference on Medical Image Computing and Computer-Assisted Intervention}, pages 525--536. Springer, 2023.

\bibitem[Liu et~al.(2023)Liu, Li, Wu, and Lee]{liu2023llava}
Haotian Liu, Chunyuan Li, Qingyang Wu, and Yong~Jae Lee.
\newblock Visual instruction tuning, 2023.

\bibitem[Liu et~al.(2021)Liu, Lin, Cao, Hu, Wei, Zhang, Lin, and Guo]{liu2021swin}
Ze Liu, Yutong Lin, Yue Cao, Han Hu, Yixuan Wei, Zheng Zhang, Stephen Lin, and Baining Guo.
\newblock Swin transformer: Hierarchical vision transformer using shifted windows.
\newblock In \emph{Proceedings of the IEEE/CVF International Conference on Computer Vision (ICCV)}, 2021.

\bibitem[Mueller and et~al.(2022)]{mueller2022lovt}
Jonas Mueller and et al.
\newblock Lovt: Local vision-text alignment for medical image analysis.
\newblock In \emph{MICCAI}, 2022.

\bibitem[Park et~al.(2022)Park, Kwon, Kim, Lee, Ha, Lim, Imani, and Kim]{park2022quiltnet}
Jongho Park, HyukJun Kwon, Seowoo Kim, Junyoung Lee, Minho Ha, Euicheol Lim, Mohsen Imani, and Yeseong Kim.
\newblock Quiltnet: Efficient deep learning inference on multi-chip accelerators using model partitioning.
\newblock In \emph{Proceedings of the 59th ACM/IEEE Design Automation Conference}, pages 1159--1164, 2022.

\bibitem[Pedraza et~al.(2015)Pedraza, Vargas, Narv{\'a}ez, Dur{\'a}n, Mu{\~n}oz, and Romero]{pedraza2015open}
Lina Pedraza, Carlos Vargas, Fabi{\'a}n Narv{\'a}ez, Oscar Dur{\'a}n, Emma Mu{\~n}oz, and Eduardo Romero.
\newblock An open access thyroid ultrasound image database.
\newblock In \emph{10th International symposium on medical information processing and analysis}, pages 188--193. SPIE, 2015.

\bibitem[Radford et~al.(2021)Radford, Kim, Hallacy, Ramesh, et~al.]{radford2021clip}
Alec Radford, Jong~Wook Kim, Jack Hallacy, Aditya Ramesh, et~al.
\newblock Learning transferable visual models from natural language supervision.
\newblock In \emph{Proceedings of the International Conference on Machine Learning (ICML)}, 2021.

\bibitem[Shazeer(2017)]{shazeer2017outrageously}
Noam et~al. Shazeer.
\newblock Outrageously large neural networks: The sparsely-gated mixture-of-experts layer.
\newblock In \emph{ICLR}, 2017.

\bibitem[Shui(2025)]{shui2025fv}
Yunhao et~al. Shui.
\newblock fvlm: Anatomy-aware vision-language model for volumetric medical imaging.
\newblock \emph{arXiv preprint arXiv:2503.00000}, 2025.

\bibitem[Singhal et~al.(2023)Singhal, Azizi, Tu, Mahdavi, Wei, Chung, Scales, Tanwani, Cole-Lewis, Pfohl, et~al.]{singhal2023large}
Karan Singhal, Shekoofeh Azizi, Tao Tu, S~Sara Mahdavi, Jason Wei, Hyung~Won Chung, Nathan Scales, Ajay Tanwani, Heather Cole-Lewis, Stephen Pfohl, et~al.
\newblock Large language models encode clinical knowledge.
\newblock \emph{Nature}, 620\penalty0 (7972):\penalty0 172--180, 2023.

\bibitem[Wang et~al.(2022)Wang, Wu, Agarwal, and Sun]{wang-etal-2022-medclip}
Zifeng Wang, Zhenbang Wu, Dinesh Agarwal, and Jimeng Sun.
\newblock Medclip: Contrastive learning from unpaired medical images and text.
\newblock In \emph{Proceedings of the 2022 Conference on Empirical Methods in Natural Language Processing}, pages 3876--3887, Abu Dhabi, United Arab Emirates, 2022. Association for Computational Linguistics.

\bibitem[Wu et~al.(2024{\natexlab{a}})Wu, Zhang, Zhang, Zhou, Zhou, and Fu]{wu2024mm}
Ruiqi Wu, Chenran Zhang, Jianle Zhang, Yi Zhou, Tao Zhou, and Huazhu Fu.
\newblock Mm-retinal: Knowledge-enhanced foundational pretraining with fundus image-text expertise.
\newblock In \emph{International Conference on Medical Image Computing and Computer-Assisted Intervention}, pages 722--732. Springer, 2024{\natexlab{a}}.

\bibitem[Wu et~al.(2024{\natexlab{b}})Wu, Chen, Pan, Liu, Liu, Dai, Gao, Ma, Wu, Wang, Xie, Wu, Hu, Wang, Sun, Li, Piao, Guan, Liu, Xie, You, Dong, Yu, Zhang, Zhao, Wang, and Ruan]{wu2024deepseekvl2mixtureofexpertsvisionlanguagemodels}
Zhiyu Wu, Xiaokang Chen, Zizheng Pan, Xingchao Liu, Wen Liu, Damai Dai, Huazuo Gao, Yiyang Ma, Chengyue Wu, Bingxuan Wang, Zhenda Xie, Yu Wu, Kai Hu, Jiawei Wang, Yaofeng Sun, Yukun Li, Yishi Piao, Kang Guan, Aixin Liu, Xin Xie, Yuxiang You, Kai Dong, Xingkai Yu, Haowei Zhang, Liang Zhao, Yisong Wang, and Chong Ruan.
\newblock Deepseek-vl2: Mixture-of-experts vision-language models for advanced multimodal understanding, 2024{\natexlab{b}}.

\bibitem[Zhang et~al.(2023)Zhang, Xu, Usuyama, Xu, Bagga, Tinn, et~al.]{zhang2023biomedclip}
Sheng Zhang, Yanbo Xu, Naoto Usuyama, Hanwen Xu, Jaspreet Bagga, Robert Tinn, et~al.
\newblock Biomedclip: A multimodal biomedical foundation model pretrained from fifteen million scientific image-text pairs.
\newblock \emph{arXiv preprint arXiv:2303.00915}, 2023.

\bibitem[Zhao et~al.(2025)Zhao, Liu, Wu, et~al.]{zhao2025clipsurvey}
Zihao Zhao, Yuxiao Liu, Han Wu, et~al.
\newblock Clip in medical imaging: A survey.
\newblock \emph{arXiv preprint arXiv:2312.07353}, 2025.

\bibitem[Zhou et~al.(2022)Zhou, Lei, Liu, Du, Huang, Zhao, Dai, Chen, Le, and Laudon]{zhou2022mixture}
Yanqi Zhou, Tao Lei, Hanxiao Liu, Nan Du, Yanping Huang, Vincent~Y. Zhao, Andrew Dai, Zhifeng Chen, Quoc Le, and James Laudon.
\newblock Mixture-of-experts with expert choice routing.
\newblock In \emph{Advances in Neural Information Processing Systems (NeurIPS)}, 2022.

\bibitem[Zhou et~al.(2023)Zhou, Zhou, Wang, et~al.]{zhou2023mrm}
Zongyu Zhou, Yuxuan Zhou, Yifan Wang, et~al.
\newblock Advancing radiograph representation learning with masked record modeling.
\newblock \emph{ICLR}, 2023.

\end{thebibliography}
}


\end{document}